\documentclass[10pt,twocolumn,letterpaper]{article}

\usepackage{toolkit/3dv}
\usepackage{times}
\usepackage{epsfig}
\usepackage{graphicx}
\usepackage{amsmath}
\usepackage{amssymb}
\usepackage{textcomp}
\usepackage{gensymb}
\usepackage{booktabs}
\usepackage{makecell}
\usepackage{enumitem}
\usepackage{subcaption}
\usepackage[ruled,vlined]{algorithm2e}
\usepackage[normalem]{ulem} 

\SetCommentSty{mycommfont}

\usepackage[pagebackref=true,breaklinks=true,letterpaper=true,colorlinks,bookmarks=false]{hyperref}
\usepackage[capitalise]{cleveref}

\threedvfinalcopy 

\def\threedvPaperID{224} 

\ifthreedvfinal\fi
\begin{document}

\title{Direct-PoseNet: Absolute Pose Regression with Photometric Consistency}

\author{Shuai Chen \qquad Zirui Wang \qquad Victor Prisacariu \\
Active Vision Lab, University of Oxford\\
{\tt\small \{shuaic, ryan, victor\}@robots.ox.ac.uk}
}


\maketitle
\ifthreedvfinal\thispagestyle{empty}\fi

\begin{abstract}
We present a relocalization pipeline, which combines an absolute pose regression (APR) network with a novel view synthesis based direct matching module, offering superior accuracy while maintaining low inference time. Our contribution is twofold: i) we design a direct matching module that supplies a photometric supervision signal to refine the pose regression network via differentiable rendering; ii) we show that our method can easily cope with additional unlabeled data without the need for external supervision such as traditional visual odometry or pose graph optimization. As a result, our method achieves state-of-the-art performance among all other single-image APR methods on the 7-Scenes benchmark and the LLFF dataset.
\end{abstract}

\section{Introduction}
Camera localization is a classical problem in computer vision and robotics. It is a core component for many applications such as virtual and augmented reality, indoor navigation systems, and autonomous driving. A typical visual-based localization algorithm is designed to determine the camera's 6-DOF positions and orientations from taking as input an RGB or RGB-D image.

The classical approach to solve this problem is built upon finding 2D-3D correspondences~\cite{Brachmann17, Brachmann18, brachmann2020dsacstar, Sattler12, Sattler17, Shotton13, Taira18} between 2D image position and 3D points in space. Then an $n$-point pose (PnP) solver is applied to the 2D-3D matches inside a RANSAC~\cite{brachmann2019ngransac, Chum08, Fischler81, Raguram13} loop. Traditionally, 2D-3D matches can be found using local feature descriptor matching, and many approaches require depth or structure-from-motion (SfM) reconstruction to build robust 3D geometric correlations~\cite{Sattler11, Sattler17}. Recent methods use machine learning to regress 3D scene coordinates from image patches~\cite{Brachmann17, Brachmann18, brachmann2020dsacstar, Shotton13} directly. Overall, 3D structure-based methods still achieve state-of-the-art (SOTA) accuracy, as discussed by Sattler \etal~\cite{Sattler19}. However, the presence of highly accurate depth images or SfM models is not universally available in real-life applications, especially for many consumer-grade devices such as smartphones or tablets. Most deep 3D structure-based methods are computation resources intensive and cannot easily achieve real-time inference with SOTA accuracy constraints.

Another line of approaches is deep learning-based pose regression~\cite{Brahmbhatt18, Kendall16, Kendall17, Kendall15, Melekhov17, Naseer17, Radwan18, Walch17, Valada18, Wu17}, also known as absolute pose regression (APR). These approaches propose to train a scene-specific deep neural network to predict 6-DoF camera pose relative to a scene directly from images. Despite obtaining inferior performances in localization benchmarks, it has gained popularity due to its high efficiency and simplicity by learning the full localization pipeline in a Convolutional Neural Network (CNN). The end-to-end approach has several appealing features compared to 3D structure-based methods: (1) most APR algorithms display great portability for commercial deployment at applications where fast and reliable performance is crucial. For example, the groundwork PoseNet~\cite{Kendall15} runs its entire process in less than $6ms$. (2) the CNN only requires RGB images input and does not rely on depth maps or SfM reconstructions, which is less hardware constrained. (3) it keeps a low memory footprint in megabytes regardless of scene sizes.

Despite these benefits, drawbacks of the APR method are also apparent. It is known to be prone to overfit the training set and significantly less accurate than structure-based methods, as shown in Sattler \etal~\cite{Sattler19} and Shavit \etal~\cite{Shavit19} . Both studies suggest that scene geometry is key for obtaining accurate pose estimation. Prior efforts have tried to add geometric constraints by finding relative pose~\cite{Balntas18, Brahmbhatt18, Radwan18, Valada18} or using reprojection error~\cite{Kendall17}. Nevertheless, it is clear that existing single image APR solutions are not yet able to compete with structure-based methods.

We address the problem of single-image APR by introducing direct matching supervision inspired by direct Visual Odometry (VO) approaches~\cite{Engel17, Engel13}. The key intuition is that the predicted pose error is inversely proportional to the visual similarity between the query image and a rendering of the 3D scene of the relocalized pose. The proposed APR framework improves pose regression using direct matching supervised by the photometric similarity between the input query image and the rendered image of the scene using the predicted pose. In the testing stage, our method runs like a standard APR method without extra computational cost. To our knowledge, this paper is the first camera pose regression method to use direct matching/photometric supervision. We summarize our contributions as follows:

\begin{itemize}[leftmargin=*]
    \item We introduce a novel camera relocalization pipeline consisting of a pose regression network and a direct matching module such that network learning is supervised by not only the traditional pose regression loss, but also a photometric loss.
    
    \item We show how unlabeled images can be leveraged using photometric loss in the direct matching module to further improve the pose regression performance without extra supervision, such as relative pose constraints.
\end{itemize}
With contributions above, our method achieves state-of-the-art performance in single-image APR on 7-Scenes benchmark and LLFF dataset.

This paper is organized as follows: we introduce existing APR methods and other related work in \cref{sec:related_work}. Our relocalization pipeline is detailed in \cref{sec:method}, with experimental results and analysis discussed in \cref{sec:exps}. \cref{sec:conclusion} concludes our work. 

    

\begin{figure*}[!htp]
    \setlength{\abovecaptionskip}{-0.2cm}
   \begin{center}
   \includegraphics[width=0.9\textwidth]{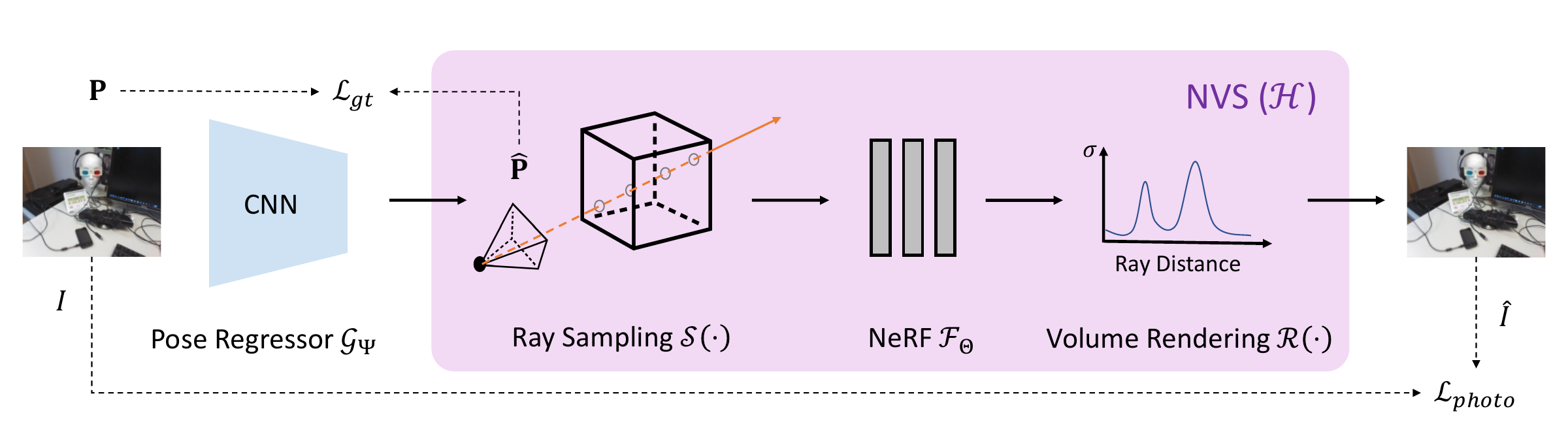}
   \end{center}
   \caption{Overview of our proposed training pipeline. Given an input image $I$, the pose regressor $\mathcal{G}_\Psi$ predicts a pose estimation $\hat{\mathbf{P}}$, from which the NVS system $\mathcal{H}$ renders a synthetic image $\hat{I}$, supplying our direct matching supervision signal $\mathcal{L}_{photo}$ and refining $\mathcal{G}_\Psi$ along with the ground truth supervision $\mathcal{L}_{gt}$.}
   \label{fig:PNeRFNet}
\end{figure*}
\section{Related Work} \label{sec:related_work}

\paragraph{Absolute Pose Regression}
Absolute pose regression methods typically require a CNN classifier that has been pre-trained from the image classification dataset. It then uses transfer learning to fine-tune the feature extractors to regress the camera pose from one or more given image sequences. To get a thorough review in this area, the interested reads is refered to Sattler \etal~\cite{Sattler19}.

The common practice in this area is introduced by PoseNet~\cite{Kendall15}. A simple pose regressor can take an arbitrary RGB image as the input and learn to regress the correspondent camera position and orientation. Successors of PoseNet focus to improve the framework in several aspects. ~\cite{Melekhov17, Walch17, Wu17} seek to enhance network architectures. LSTM PoseNet~\cite{Walch17} combines LSTM with CNN to reduce feature dimensions for pose regression. Hourglass PoseNet~\cite{Melekhov17} adapts an encoder-decoder style backbone. BranchNet~\cite{Wu17} uses a multi-task CNN where low-level common features are extracted before splitting the network into two separate branches to predict the camera position and orientation. Kendall and Cipolla~\cite{Kendall17} proposed learnable weights to sidestep hyperparameter tunning that balances the translation and rotation loss in PoseNet. Besides learning the optimal weight between losses, they attempt to leverage scene geometry for pose regression by formulating a reprojection error between the ground-truth pose and predicted pose. While we share the same insight that geometry would help pose regression, we differ from theirs in that 1) the scene geometry is implicitly represented in a novel view synthesis-enabled direct matching module; 2) we introduce a dense pixel-level photometric supervision.

\paragraph{Unlabeled training in APR}
Rather than only training on images with ground-truth pose annotation, MapNet~\cite{Brahmbhatt18} is able to train on unlabeled video frames by adding pair-wise geometric constraints between video frames using additional VO algorithms~\cite{Engel17, Engel13}. During inference, it can utilize pose graph optimization (PGO) for post-processing to further boost the performance.
We also recognize the importance of using additional unlabeled data in this work. 
However, instead of acquiring constraints from an additional VO algorithm or PGO, which assumes video frames are available, our method can train on images from arbitrary viewpoints by minimizing the photometric loss between the rendered images and the input images.

\paragraph{Direct Matching for Motion Estimation}
Direct matching methods or direct methods refer to the commonly used methods to recover camera motions by directly measuring image intensities~\cite{Irani99} in VO and simultaneous localization and mapping (SLAM) systems. In contrast to the feature-based methods~\cite{Davison07, Soatto00, Klein07, Mur-Artal15} that minimize reprojection errors based on corresponding features among frames, direct methods exploit all information in the image and recover camera motions by minimizing the photometric error. Compared to feature-based methods, direct methods are often more reliable in sparse textured environments and do not have feature extraction operations to add in computation costs. DTAM~\cite{Newcombe11}, REMODE~\cite{Pizzoli14}, and LSD-SLAM~\cite{Engel13} employ dense reconstruction using direct methods. SVO~\cite{Forster14, Forster17} proposes a hybrid approach to implement direct VO motion estimation at the SLAM frontend and uses feature-based methods in the backend mapping thread. DSO~\cite{Engel17} further proposes a sparse and direct method. This paper is partly inspired by direct methods and adopts photometric supervision in training the pose regression network. However, our method is different from direct VO methods in several aspects. First, our method does not compute photometric errors based on image intensities but RGB differences. Second, our method provides absolute pose estimation using a single image, but the methods above are designed to take a pair of neighboring frames for computing relative motion.

\paragraph{Novel View Synthesis}
Novel view synthesis (NVS) is a long-standing problem in computer graphics. It aims to generate novel camera perspectives based on image samples of the scenes~\cite{Tewari20}. Early works in this field can be traced back to nearly 30 years ago when some required using densely captured views of the scene~\cite{Gortler96, Levoy96}, and others~\cite{Chen93} interpolate novel views using image warping. Recently, novel view synthesis has made rapid progress in achieving photorealistic view synthesis~\cite{liu2020neural, Lombardi19, martinbrualla2020nerfw, Meshry19, Mildenhall20, park2020nerfies, Dai20, Schwarz2020NEURIPS, Sitzmann19, Thies19} with sparser view samples, thanks to recent development in neural 3D shape representation~\cite{Mescheder19, Niemeyer20, Park19, Sitzmann19SRN}. We select a recent popular approach from Mildenhall \etal~\cite{Mildenhall20} to build our camera re-localization training pipeline in this work. Specifically, we incorporate NeRF architecture to provide photometric supervision to the pose regression model. 
We consider two other NeRF-based works that are able to estimate camera pose related to our paper: iNeRF~\cite{yen2020inerf} estimates pose iteratively by inverting a pre-trained NeRF model on the test images. Wang~\etal~\cite{wang2021nerfmm} show that 3D scene representation and camera poses can be jointly optimized within a NeRF framework. 
Nonetheless, the fundamental difference of the proposed method to theirs is that we only use differentiable rendering to compute photometric loss in training of our pose regressor. After training, we are able to predict camera pose in a single network forward pass, whereas their methods require iterative optimization in test time.
\section{Method} \label{sec:method}
\cref{fig:PNeRFNet} illustrates our proposed relocalization pipeline, which consists of an NVS-enabled direct matching module and a pose regression network. We aim to predict a camera pose $\hat{v}$ for an input image $I$ via a pose regression network $\mathcal{G}_\Psi$, which is supervised by a novel direct matching signal along with ground truth poses during training. At test time, only the pose regression network is required, ensuring rapid inference while offering superior relocalization accuracy.

This section is organized as follows: \cref{sec:direct_matching} details our direct matching module. 
A full system setup is explained in \cref{sec:system_setup}. To explore the possibility of utilizing more data, a further unlabeled training scheme is explained in \cref{sec:unlabeled_training}.

\subsection{Direct Matching} \label{sec:direct_matching}
Direct matching is a common approach in traditional SLAM and VO systems~\cite{Engel17, Engel13, Forster14, Forster17, Newcombe11, Pizzoli14}. It refers to the process that optimizes camera poses via minimizing a photometric loss. In this work, we adapt the direct matching concept and apply it to assist the training of our pose regression network. Concretely, assuming a pre-trained pose regression network $\mathcal{G}_\Psi$ and an NVS system $\mathcal{H}$ are available, given an image $I$ captured at viewpoint $v$ and its pose estimation $\hat{v} = \mathcal{G}_\Psi(I)$, our direct matching module constrains $\Psi$ via minimizing the photometric difference $\mathcal{L}_{photo}(\hat{I}, I)$ between a synthetic image $\hat{I} = \mathcal{H}(\hat{v})$ rendered by the NVS model $\mathcal{H}$ at viewpoint $\hat{v}$, and its true observation $I$:
\begin{equation}
    \mathcal{L}_{photo}(\hat{I}, I) = \| \hat{I} - I\|_2.
\label{eq:photo_loss}
\end{equation}
Any NVS system with a differentiable renderer could be selected as the NVS system $\mathcal{H}$. For simplicity, we choose the NeRF \cite{Mildenhall20} in this work, due to the high quality reconstructions it produces. 

From a high-level point of view, NeRF-based NVS requires three main components: 1) a neural radiance field function $\mathcal{F}_\Theta$ that models a 3D volume, 2) a differentiable volume renderer $\mathcal{R}(\cdot)$ that enables back-propagation, and 3) a viewpoint-dependent volume sampling function $\mathcal{S}(\cdot)$ that provides 3D sample locations and view directions for $\mathcal{F}_\Theta$ and $\mathcal{R}(\cdot)$ given a camera pose. Consequently, an image rendered from an NVS system $\mathcal{H}$ using NeRF can be formulated by:
\begin{equation}
    \hat{I}_{v} = \mathcal{H}(v) \triangleq \mathcal{R}(\mathcal{F}_\Theta, \mathcal{S}(v)),
\end{equation}
where $\hat{I}_{v}$ denotes a synthetic image rendered at a pose $v$, and all operations above are differentiable. As a result, our direct matching system updates $\Psi$ by minimizing photometric loss in \cref{eq:photo_loss}.

\subsection{System Setup} \label{sec:system_setup}
\paragraph{Training Pipeline}
As mentioned above, our system consists of two main components, a pose regression network $\mathcal{G}_\Psi$ and an NVS-enabled direct matching module. With each component pre-trained on target data, we join them together to further refine the pose regression network: given an input image $I$, the pose regression network predicts a pose $\hat{v}$, from which an NVS system $\mathcal{H}$ renders a synthetic image $\hat{I}_{\hat{v}} = \mathcal{H}(\hat{v})$, enabling our direct matching supervision $\mathcal{L}_{photo}$. Meanwhile, the ground truth supervision $\mathcal{L}_{gt}$ is applied as well. As a result, the pose regression network $\mathcal{G}_\Psi$ is refined through a back-propagation of a weighted sum of $\mathcal{L}_{photo}$ and $\mathcal{L}_{gt}$ ( \cref{eq:refine_posenet}).
Mathematically, our training pipeline can be framed as:
\begin{equation}
    \Psi^* = \arg\min_{\Psi}(\lambda_1\mathcal{L}_{photo} + \lambda_2\mathcal{L}_{gt}),
    \label{eq:refine_posenet}
\end{equation}
where $\Psi^*$ denotes optimized network parameters, and $\lambda_1, \lambda_2$ denote weights for each loss terms, respectively.

\paragraph{Pose Regression Network}
Our pose regression network $\mathcal{G}_\Psi$ follows the line of PoseNet \cite{Kendall15} work, including a pre-trained feature extractor backbone and a fully connected layer that outputs a camera pose matrix $\hat{\mathbf{P}}$. 
Prior works \cite{Brahmbhatt18, Kendall16, Kendall17, Kendall15, Melekhov17, Radwan18, Valada18, Walch17} typically use a quaternion or an axis-angle representation during pose estimation, requiring balancing between rotation and translation terms. Instead, our network regress a camera pose $v$ using the representation of $\mathbf{P} = [\mathbf{R}|\mathbf{t}]$ to overcome this issue, where $\mathbf{R} \in \text{SO}(3)$ is a rotation matrix denotes a camera orientation and $\mathbf{t} \in \mathbb{R}^{3}$ denotes camera position. For clarity, we refer $v$ as a general pose concept and $\mathbf{P}$ as a specific pose representation.

Our ground truth supervision loss is defined as the L2 distance between a ground truth pose $\mathbf{P}$ and an estimated pose $\hat{\mathbf{P}}$:
\begin{equation} \label{eq:L_gt}
  \mathcal{L}_{gt} = \|\mathbf{P}-\hat{\mathbf{P}}\|_{2},
\end{equation}
removing the need for balancing rotation and translation terms while offering competitive performance.

\paragraph{NeRF}
Two challenges arise while adapting NeRF as our NVS system $\mathcal{H}$ in relocalization context. First, the training cost is expensive in relocalization tasks, where a training video could easily yield thousands of images even after frame subsampling. We resolve this issue by reducing the NeRF model size and removing the hierarchical training scheme.

Second, strong artifacts occur in synthetic images if the NeRF is applied in our task without modifications. Two reasons account for that: a) NeRF is not designed for outward-looking scenes \cite{kaizhang2020} and b) photometric consistency is violated when auto-focus/exposure fluctuation and rolling shutter effect appear. We mitigate this issue by adapting a coarse-to-fine positional encoding scheme $\gamma_{\alpha}(\cdot)$ from Nerfies \cite{park2020nerfies}. Specifically, an input signal $p$ is encoded by $\gamma_{\alpha}(p) = [p, \dots, w_k(\alpha_t) \sin (2^{k} \pi p), w_k(\alpha_t) \cos (2^{k} \pi p), \dots]$, where $0 \leq k \leq m-1, m\in\mathbb{N}$ and $w_{k}(\alpha_t)$ activates each band over epoch $t$, controlled by $\alpha_t = mt/N$. $N$ is a user-defined maximum epoch number in training, where $k$ reaches the maximum frequency band $m-1$. We refer interested readers to the full mathematical expression for $w_k(\alpha_t)$ in our supplementary material.

With the progressive positional encoding function, we are able to adapt NeRF as our NVS system $\mathcal{H}$, reducing artifacts and preserving high frequency details. Please refer to \cref{sec:ablation_study} for more discussion on the effectiveness in this approach. 

\subsection{Unlabeled Training} \label{sec:unlabeled_training}
Inspired by MapNet+ \cite{Brahmbhatt18}, we propose to improve pose estimation in a semi-supervised manner, with unlabeled sequences captured in the same training scene. Unlike \cite{Brahmbhatt18}, which enforces a relative geometric constraint between two nearby frames and requires an additional VO algorithm, we rely on our bootstrapped pipeline to further refine the pose regression network $\mathcal{G}_\Psi$.
Given an input image without ground truth pose annotation but not too far from labeled training videos, 
the training of $\mathcal{G}_\Psi$ can be supervised by the photometric loss between the synthetic image rendered by the direct matching module using the predicted pose. This semi-supervised training scheme can be effectively set up by setting $\lambda_1=1.0$ and $\lambda_2=0.0$. 
We find our unlabeled training works well, evidenced by the performance in \cref{table:2}.

\begin{table*}[]
   \resizebox{\textwidth}{!}{%
      \begin{tabular}{lccccccccc|ccc}
         \toprule
         &
         \multicolumn{9}{c|}{without unlabeled data} &
         \multicolumn{3}{c}{with unlabeled data} \\
         \midrule
         Scene & \makecell[c]{PN~\cite{Kendall15}} &
         \makecell[c]{PN learned \\weights~\cite{Kendall17}} &
         \makecell[c]{geo. PN\\~\cite{Kendall17}} &
         \makecell[c]{LSTM PN\\~\cite{Walch17}} &
         \makecell[c]{Hourglass\\PN~\cite{Melekhov17}} &
         \makecell[c]{BranchNet\\~\cite{Wu17}} &
         \makecell[c]{DSO\\~\cite{Engel17}} &
         \makecell[c]{MapNet\\~\cite{Brahmbhatt18}} &
         \makecell[c]{\textcolor{red}{Direct-PN}} &
         \makecell[c]{MapNet+\\~\cite{Brahmbhatt18}} &
         \makecell[c]{MapNet+\\PGO~\cite{Brahmbhatt18}} &
         \makecell[c]{\textcolor{red}{Direct-PN+U}} \\
         \midrule
         Chess   & 0.32/8.12 & 0.14/4.50 & 0.13/4.48 & 0.24/5.77 & 0.15/6.17 & 0.18/5.17 & 0.17/8.13 & \textcolor{red}{0.08}/\textcolor{red}{3.25} & 0.10/3.52 & 0.10/3.17 & \textcolor{red}{0.09}/3.24 & \textcolor{red}{0.09}/\textcolor{red}{2.77} \\
         Fire    & 0.47/14.4 & 0.27/11.8 & 0.27/11.3 & 0.34/11.9 & 0.27/10.8 & 0.34/8.99 & \textcolor{red}{0.19}/65.0 & 0.27/11.7 & 0.27/\textcolor{red}{8.66} & 0.20/9.04 & 0.20/9.29 & \textcolor{red}{0.16}/\textcolor{red}{4.87} \\
         Heads   & 0.29/12.0 & 0.18/12.1 & \textcolor{red}{0.17}/13.0 & 0.21/13.7 & 0.19/\textcolor{red}{11.6} & 0.20/14.2 & 0.61/68.2 & 0.18/13.3 & \textcolor{red}{0.17}/13.1 & 0.13/11.1 & 0.12/8.45 & \textcolor{red}{0.10}/\textcolor{red}{6.64} \\
         Office  & 0.48/7.68 & 0.20/5.77 & 0.19/5.55 & 0.30/8.08 & 0.21/8.48 & 0.30/7.05 & 1.51/16.8 & 0.17/\textcolor{red}{5.15} & \textcolor{red}{0.16}/5.96 & 0.18/5.38 & 0.19/5.42 & \textcolor{red}{0.17}/\textcolor{red}{5.04} \\
         Pumpkin & 0.47/8.42 & 0.25/4.82 & 0.26/4.75 & 0.33/7.00 & 0.25/7.0  & 0.27/5.10 & 0.61/15.8 & 0.22/4.02 & \textcolor{red}{0.19}/\textcolor{red}{3.85} & \textcolor{red}{0.19}/3.92 & \textcolor{red}{0.19}/3.96 & \textcolor{red}{0.19}/\textcolor{red}{3.59} \\
         Kitchen & 0.59/8.64 & 0.24/5.52 & 0.23/5.35 & 0.37/8.83 & 0.27/10.2 & 0.33/7.40 & 0.23/10.9 & 0.23/\textcolor{red}{4.93} & \textcolor{red}{0.22}/5.13 & 0.20/5.01 & 0.20/4.94  & \textcolor{red}{0.19}/\textcolor{red}{4.79} \\
         Stairs  & 0.47/13.8 & 0.37/10.6 & 0.35/12.4 & 0.40/13.7 & 0.29/12.5 & 0.38/\textcolor{red}{10.3} & \textcolor{red}{0.26}/21.3 & 0.30/12.1 & 0.32/10.61 & 0.30/13.4 & 0.27/10.6 & \textcolor{red}{0.24}/\textcolor{red}{8.52} \\
         \midrule
         Average &
         0.44/10.44 &
         0.24/7.87 &
         0.23/8.12 &
         0.31/9.85 &
         0.23/9.53 &
         0.29/8.30&
         0.26/29.4 &
         0.21/7.77 &
         \textcolor{red}{0.20}/\textcolor{red}{7.26} &
         0.19/7.29 &
         0.18/6.55 &
         \textcolor{red}{0.16}/\textcolor{red}{5.17}\\
         \bottomrule
         \\
      \end{tabular}%
   }
   \caption{Pose regression results on 7 Scenes datasets. We compare our method with both direct matching and absolute pose regression methods, in median translation error (m) and rotation error (\degree). Bottom row is the average of median errors of all scenes. PN denotes PoseNet. Numbers in \textcolor{red}{red} represent the best performance with or without unlabeled data.}
   \label{table:1}
\end{table*}
\section{Experiments} \label{sec:exps}
In the following, we discuss the implementation details of our solution in~\cref{sec:4.1}. We perform a thorough evaluation of the proposed method in~\cref{sec:4.2} on the 7-Scenes dataset. We further evaluate our method on the LLFF dataset to demonstrate that the proposed method benefits from both the traditional pose regression method and the direct matching method (\cref{Sec:LLFF}). Finally, to gain more insights to our modification on positional encoding and the effectiveness of direct matching for camera localization, we apply more experiments in the ablation study (\cref{sec:ablation_study}).

\subsection{Implementation Details} \label{sec:4.1}
\paragraph{Pose Regression}
We build our pose regression model upon prior DNN-based methods~\cite{Brahmbhatt18, Kendall17, Kendall15} using PyTorch~\cite{NEURIPS2019_9015}. We choose to use the MobileNetV2~\cite{MobileNetv2} backbone in this work. We freeze the batch normalization layers~\cite{Ioffe15} from the pre-trained ImageNet backbone to train our baseline pose regression model. Since a direct rotation matrix regression may not belong to SO(3), a singular value decomposition (SVD) is applied to normalize the rotation component of $\mathbf{\hat{P}}$ during inference time. However, we also find that the pose regression network learns to predict orthogonal rotation matrices even without using the SVD. All models are optimized with the Adam optimizer~\cite{Kingma15}. The base model is trained with a batch size of 4 and a learning rate of $1 \times 10^{-4}$. We implement an early stopping strategy with a patience value of 200 and schedule the learning rate decay for every 50 epochs on validation loss plateau with a factor of 0.95.

\paragraph{NeRF}
Our NeRF model is trained with input poses in SE(3). To ensure a consistent coordinate system in pose regression and NVS, we further align and recenter the camera poses with zero-means similar in Mildenhall \etal~\cite{Mildenhall20}. 
The NeRF architecture mainly follows the original implementation~\cite{Mildenhall20}, except we apply a coarse-to-fine positional encoding $\gamma_{\alpha}(p)$ for both positions and directions. We set the maximum frequency band $m=8$, and time to reach the maximum frequency band $N=1200$ epochs.

\paragraph{Training}
For our proposed methods in \cref{sec:system_setup} and \cref{sec:unlabeled_training}, we train the pose regression model with direct matching (Direct-PoseNet) with $\lambda_1=0.3$ and $\lambda_2=0.7$, and further fine-tune the model (Direct-PoseNet+U) with $\lambda_1=1.0$ and $\lambda_2=0.0$ to simulate the unlabeled data circumstances. We set the batch size to 1 for training both models. The learning rate is set to $1 \times 10^{-5}$ with the same early stopping strategy as above. All models are trained within 24 hours with a single Nvidia 1080Ti graphic card. In our experience, the NeRF training time can reach approximately the same as training PoseNet models. More details on network architecture and training procedure are provided in the supplementary material.

\subsection{Evaluation on 7-Scenes} \label{sec:4.2}
\begin{figure*}
   \centering
    \begin{subfigure}[b]{.41\columnwidth}
        \centering
        \includegraphics[width=1\linewidth]{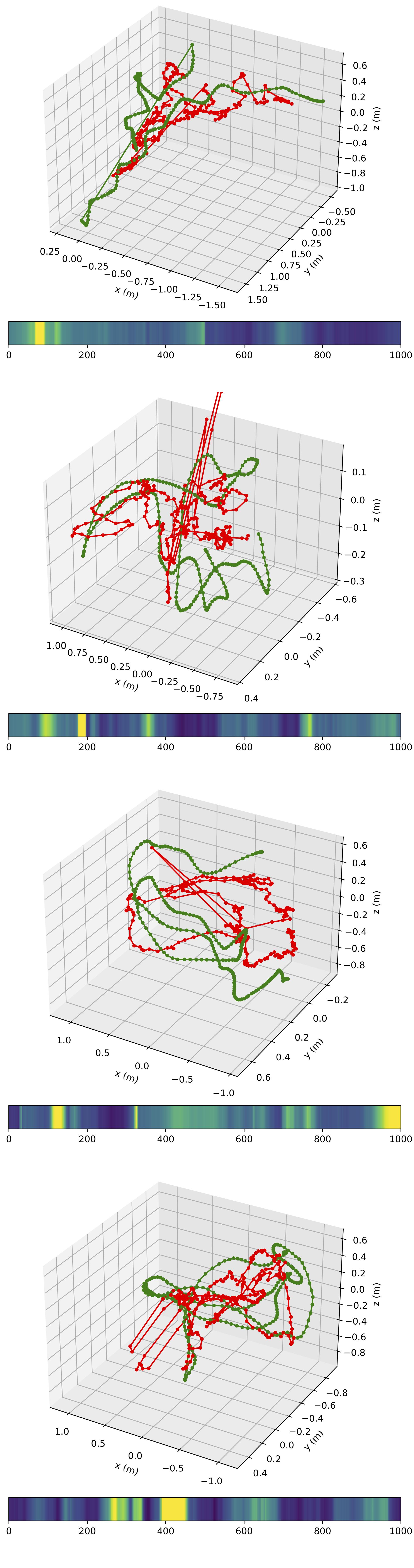}%
        \caption{PoseNet~\cite{Brahmbhatt18, Kendall17, Kendall15}}
    \end{subfigure}\begin{subfigure}[b]{.41\columnwidth}
       \centering
       \includegraphics[width=1\linewidth]{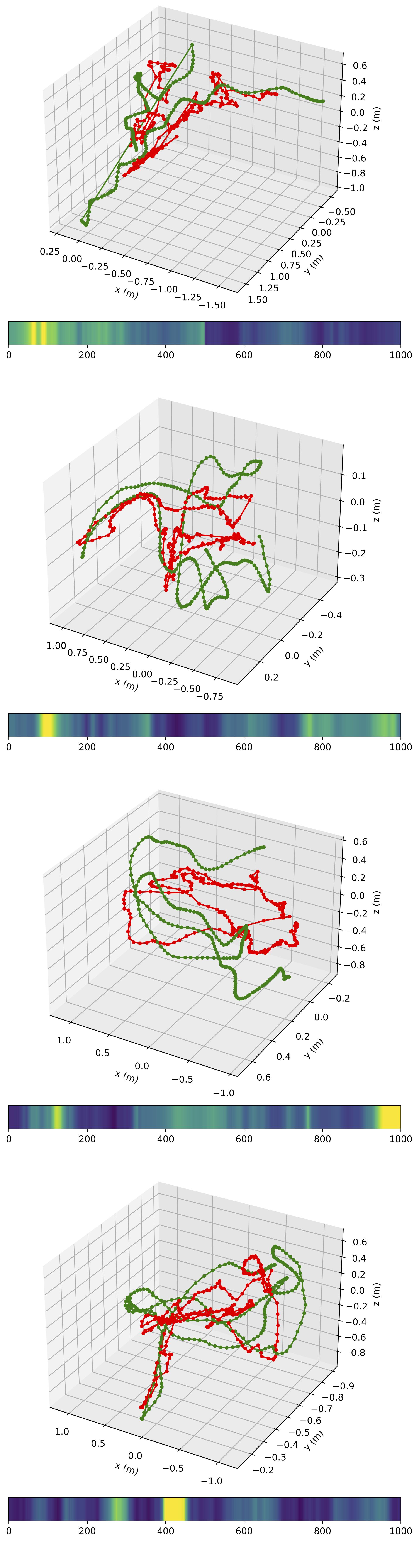}%
       \caption{MapNet~\cite{Brahmbhatt18}}
    \end{subfigure}\begin{subfigure}[b]{.41\columnwidth}
       \centering
       \includegraphics[width=1\linewidth]{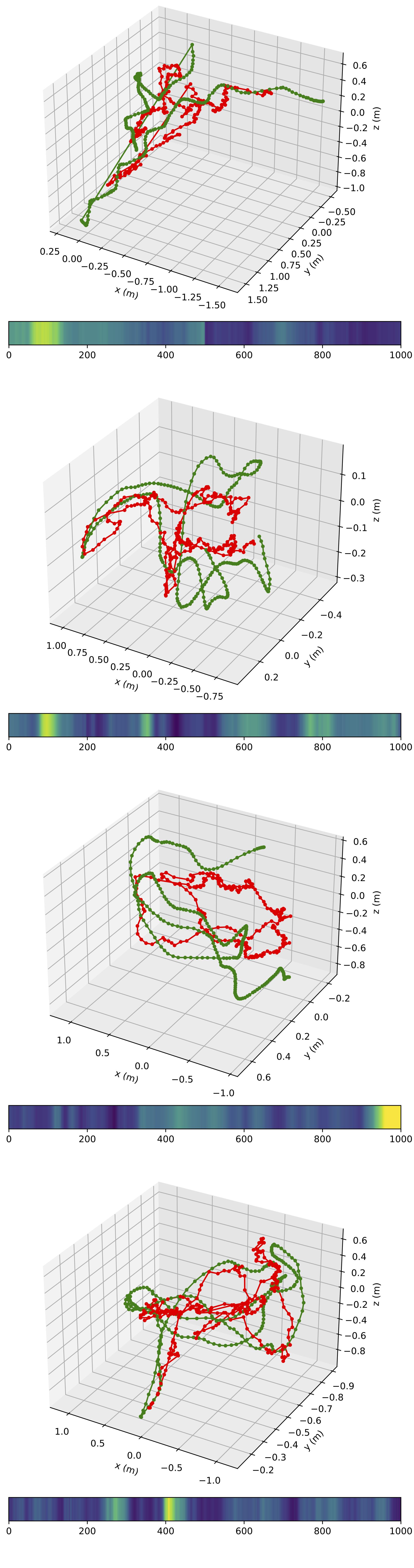}%
       \caption{Direct-PoseNet}
    \end{subfigure}\begin{subfigure}[b]{.41\columnwidth}
        \centering
        \includegraphics[width=1\linewidth]{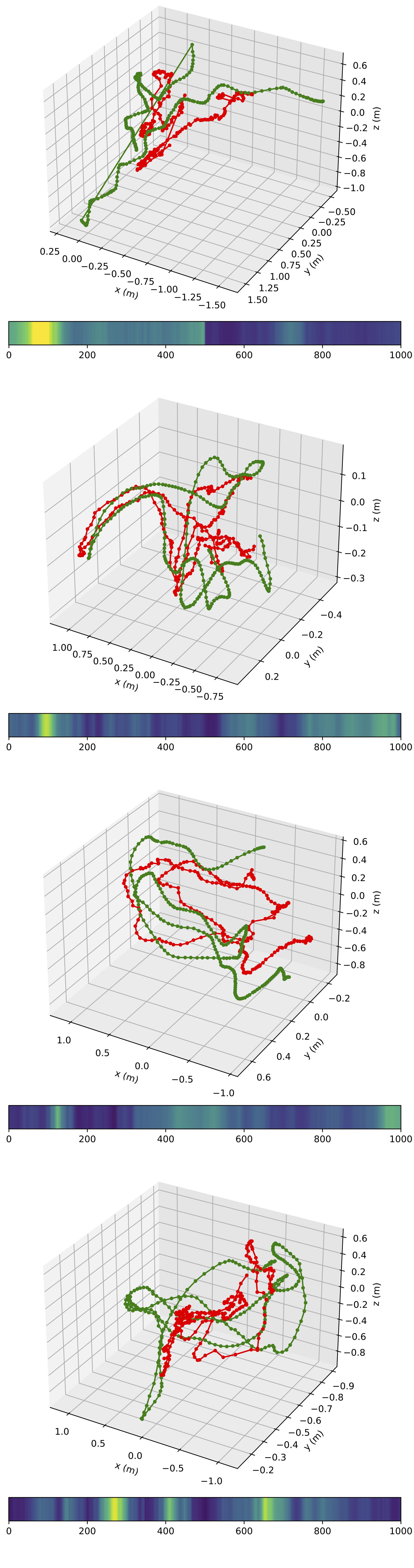}%
        \caption{MapNet+PGO~\cite{Brahmbhatt18}}
    \end{subfigure}\begin{subfigure}[b]{.41\columnwidth}
        \centering
        \includegraphics[width=1\linewidth]{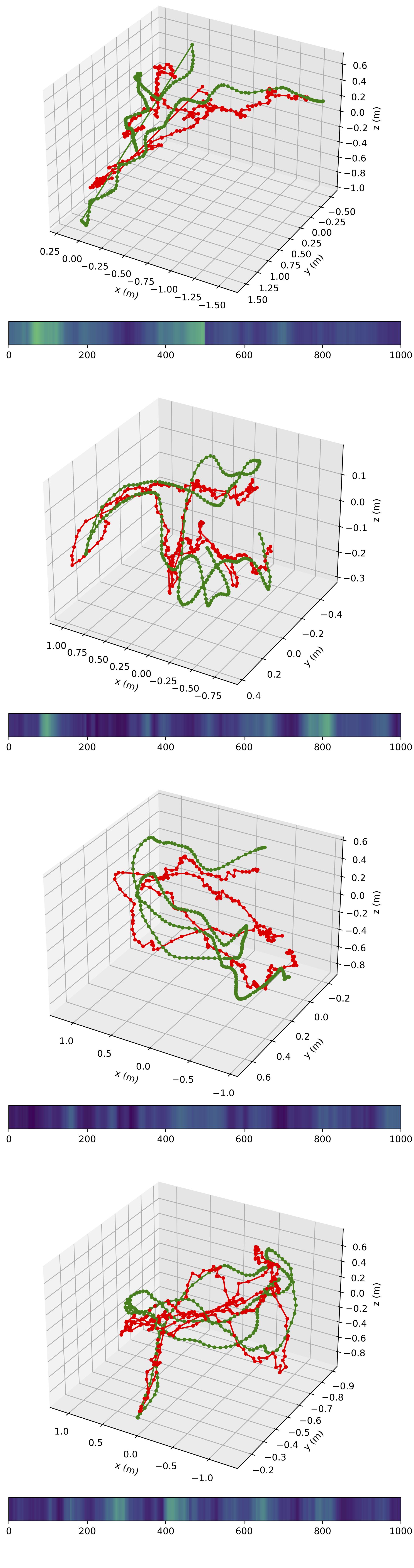}%
        \caption{Direct-PoseNet+U}
    \end{subfigure}
\caption{Visualization of camera relocalization results on 7-Scenes dataset~\cite{Glocker13, Shotton13}. For each 3D plot, we show the ground truth camera trajectory in \textcolor{green}{green} and the predicted trajectory in \textcolor{red}{red}. The bottom color bar represents rotation errors for each subplot. Yellow represents high rotation error, and blue represents low rotation error for each test sequence. Sequence names from top to bottom are: Stairs-all, Heads-all, Fire-seq-03, Office-seq-09.}
\label{fig:7Scene_visual_comparison}
\end{figure*}
We evaluate our method on a well-known camera localization dataset 7-Scenes~\cite{Glocker13, Shotton13}. It consists of seven indoor scenarios, each scale from $2m^3$ to $6m^3$. The sequences were shot by Kinect RGB-D camera at $640 \times 480$ resolution, and the ground truth poses were obtained by a dense 3D model.

The pose regression network takes an input image in $320 \times 240$ and the pre-trained NeRF model is trained with resized images in $160 \times 120$, but inference in $320 \times 240$ for Direct-PoseNet and Direct-PoseNet+U training. For each scene, we train our NeRF model with a learning rate of $5 \times 10^{-4}$ for 4000 epochs with Adam optimizer and decays exponentially to $8 \times 10^{-5}$ throughout the course of optimization. We set the near and far bounds $[b_n, b_f]$ to $[0.5, 4]$ except for the Heads scene, we set them to $[0.5, 2.5]$. However, unlike the original NeRF~\cite{Mildenhall20}, which uses a coarse-to-fine sampling approach in its architecture, we only use a single MLP model with a width of 128. We sample one image with a batch of 1024 rays for each iteration, and each ray uniformly samples $N=128$ bins. The above modifications achieve approximately $3\times$ speed up compare to the original NeRF paper. To further improve training efficiency, our NeRF model, Direct-PoseNet model, and Direct-PoseNet+U model only use a spacing window $d=5$ of the training set for scenes contain $\leq$ 2000 frames, and $d=10$ of the training set otherwise.

We summarize complete quantitative comparisons of our proposed method with prior absolute pose regression works and DSO in \cref{table:1}. For the experiment of Direct-PoseNet+U, we follow MapNet+ to use the unlabeled test sequences for fine-tuning. We do not use the entire test sequences for fine-tuning, but only 1/5 or 1/10 of the sequences described above to ensure our method is not overfitting to the entire test sequences. We also demonstrate a selection of the visual comparisons in \cref{fig:7Scene_visual_comparison}.

\subsection{Evaluation on LLFF} \label{Sec:LLFF}
We further evaluate our method on another real-world complex scene dataset, the LLFF dataset~\cite{Mildenhall15}. The dataset consisting of 8 forward-facing scenes captured with a handheld cellphone and holds out $1/8$ of the data as the test set. It is ideal for an experiment because a high-quality NeRF can be trained on LLFF to examine the combined effects of pose regression and direct matching supervision. 

\begin{table}[b]
   \resizebox{\linewidth}{!}{%
   \begin{tabular}{l|lll}
   \toprule
   \makecell[c]{\textit{error rate} in \%}  & iNeRF     & PoseNet+SE(3) & Direct-PoseNet+U                               \\
   \midrule
   \textless 5cm, \textless 5° & 73\%, 71\% & 57\%, 100\% & 78\%, 100\%\\
   \bottomrule
   \end{tabular}%
   }
   \caption{We report the percentage of correctly re-localized frames below an error threshold of 5cm and percentage of re-localized frames below an error threshold $5\degree$ on the \textit{Fern}, \textit{Fortress}, \textit{Horns}, \textit{Room} scenes of LLFF dataset~\cite{Mildenhall15}}
   \label{table:2}
\end{table}
We compare our method with both the pose regression method and an inverting NVS method iNeRF by Lin \etal~\cite{yen2020inerf}. The iNeRF uses an iterative optimization approach on each test image to recover the camera pose by inverting a trained NeRF model. On the other hand, our method does not rely on iterative optimization and produces a more generalized and efficient model. To ensure a fair comparison, we trained our NeRF model to follow the same setting with Lin \etal, which uses a standard NeRF model with a ray batch size of 2048. We fine-tune the NeRF model on four scenes (\textit{Fern}, \textit{Fortress}, \textit{Horn}, \textit{Room}) with the baseline pose regression model and compute the percentage of predicted pose whose error is less than 5cm and the percentage of predicted pose whose error is less than $5\degree$. We report the experiment results in \cref{table:2}. We observe that our proposed pose regression method gains benefit both from the pose regression approach and the direct matching approach, resulting in the top performance.
\subsection{Ablation Study}  \label{sec:ablation_study}

\paragraph{Effectiveness of Modified Positional Encoding}
In real-life datasets such as 7-Scenes, there are multiple sources to keep NeRF from rendering a high-quality, photorealistic view of the scene. Artifacts may be produced by letting the NeRF model learn from images with severe deformation (\eg, from camera rolling shutter or deforming object) or motion blur from long exposure among frames. Moreover, training and testing in very different camera trajectories is a situation for NeRF likely to fail because it tries to generate the scene from unfamiliar volumetric rendering locations.
\begin{figure}[b]
   \centering
   \begin{subfigure}[b]{.5\columnwidth}
       \centering
       \includegraphics[width=1\linewidth]{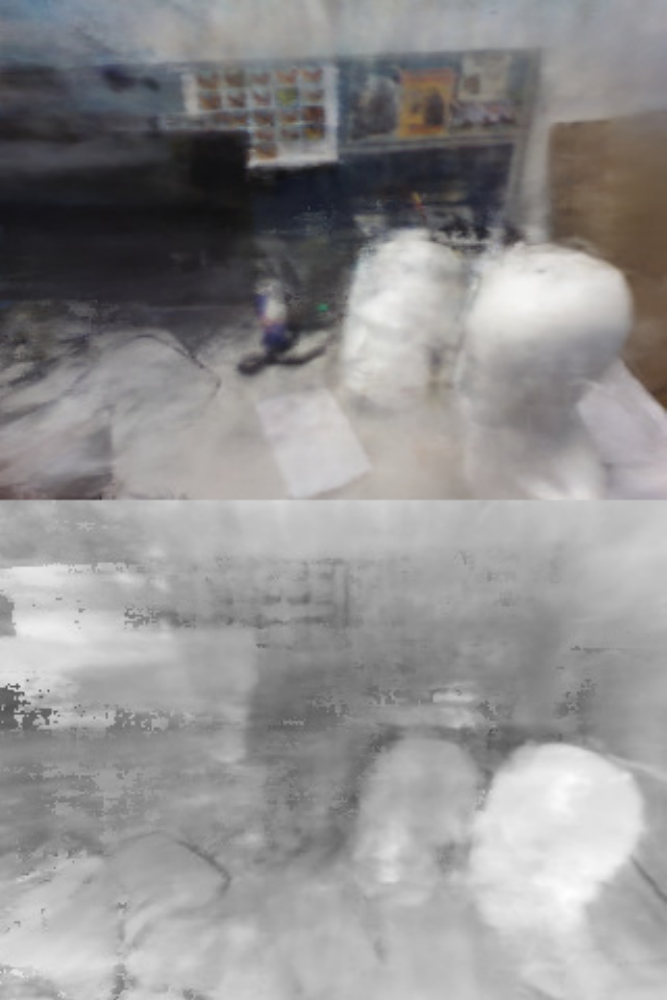}%
       \caption{Fixed P.E.~\cite{Mildenhall20}}
   \end{subfigure}\begin{subfigure}[b]{.5\columnwidth}
       \centering
       \includegraphics[width=1\linewidth]{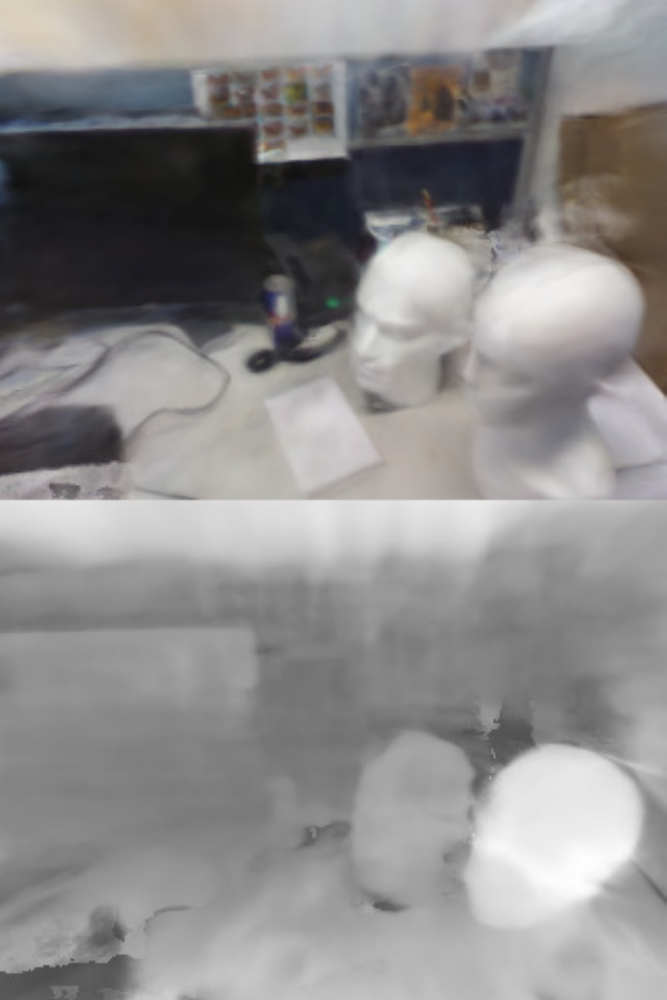}%
       \caption{Coarse-to-fine P.E.}
   \end{subfigure}
\caption{A visual comparison between (a) fixed positional encoding (P.E.) and (b) the coarse-to-fine P.E in Heads scene. Top: testset renderings from two NeRF models with different P.E. schemes. Bottom: disparity maps (inverse depth). Notice that NeRF with fixed P.E. produces stronger artifacts in this outward looking scene. Even though our encoding do not completely remove all artifacts, it recovers more structure and details than the original NeRF scheme. We provide more detailed discussion on why NeRF suffers from severe artifacts in~\cref{sec:ablation_study}.}
\label{fig:NeRFvsDNeRF}
\end{figure}

We build a toy example to demonstrate the phenomenon in the 7-Scenes dataset, and the effectiveness of our modification in NeRF's positional encoding. We randomly select a frame in Heads and sample a portion of training and validation data that lies within its frustum overlap threshold using an approach similar to Balntas \etal~\cite{Balntas18}. For this experiment, we set the frustum overlap threshold to be 0.85. We report the peak signal-to-noise ratio (PSNR) on this toy dataset for fixed full encoding ($m=10$) in the original NeRF paper, fixed half encoding ($m=5$), and the coarse-to-fine encoding schemes in~\cref{table:3}. We show a qualitative comparison between the original NeRF embedding scheme and our modified coarse-to-fine scheme in \cref{fig:NeRFvsDNeRF}. Overall, we find that using a coarse-to-fine positional encoding approach generally obtains a higher quality NeRF model throughout the 7-Scenes dataset in our experiments.
\begin{table}[h]
   \centering
   \resizebox{0.9\linewidth}{!}{%
   \begin{tabular}{l|lll}
   \toprule
   Model & \makecell[c]{Full encoding} & \makecell[c]{Half encoding} & \makecell[c]{Coarse-to-fine} \\
   \midrule
   PSNR & \makecell[c]{16.64}                      & \makecell[c]{17.16}                    & \makecell[c]{\textbf{17.50}}  \\               
   \bottomrule   
   \end{tabular}%
   }
   \caption{Comparison of different NeRF positional encoding scheme in our toy dataset (validation split).}
   \label{table:3}
\end{table}

\paragraph{Effectiveness of Direct Matching}
We investigate the effectiveness of direct matching for supervising the pose regression training. We first train a NeRF model using the Heads data. We randomly perturb the ground truth pose in different ranges and compute the photometric loss $\mathcal{L}_{ph\_perturb}$ with the ground truth image. We then count the percentage that $\mathcal{L}_{ph\_perturb} < \mathcal{L}_{ph\_GT}$, where $\mathcal{L}_{ph\_GT}$ denotes the photometric loss using the ground truth pose. We define such a percentage as the \textit{error rate}. Ideally, views generated from perturbed poses should have higher photometric loss than the loss we get from using the ground truth poses. Thus the \textit{error rate} indicates the chances that non-ideal cases would happen, which potentially damages the optimization of our pose regression.
\begin{figure}[b]
    \centering
    \captionsetup[subfigure]{position=top}
    \begin{subfigure}[b]{.5\columnwidth}
        \centering
        \includegraphics[width=1\linewidth]{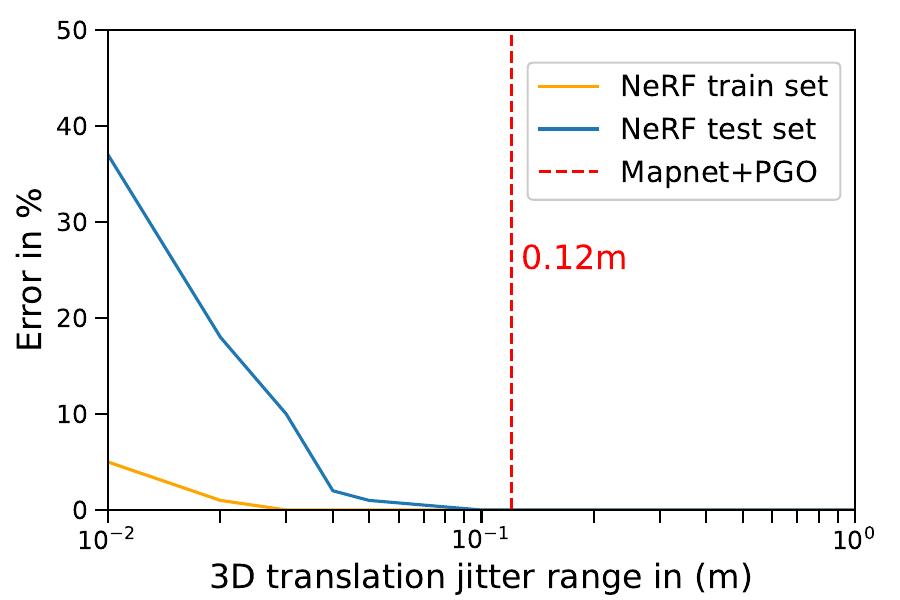}%
        \captionsetup{position=top}
        \subcaption{\small 3D translation jitter only}
    \end{subfigure}\begin{subfigure}[b]{.5\columnwidth}
        \centering
        \includegraphics[width=1\linewidth]{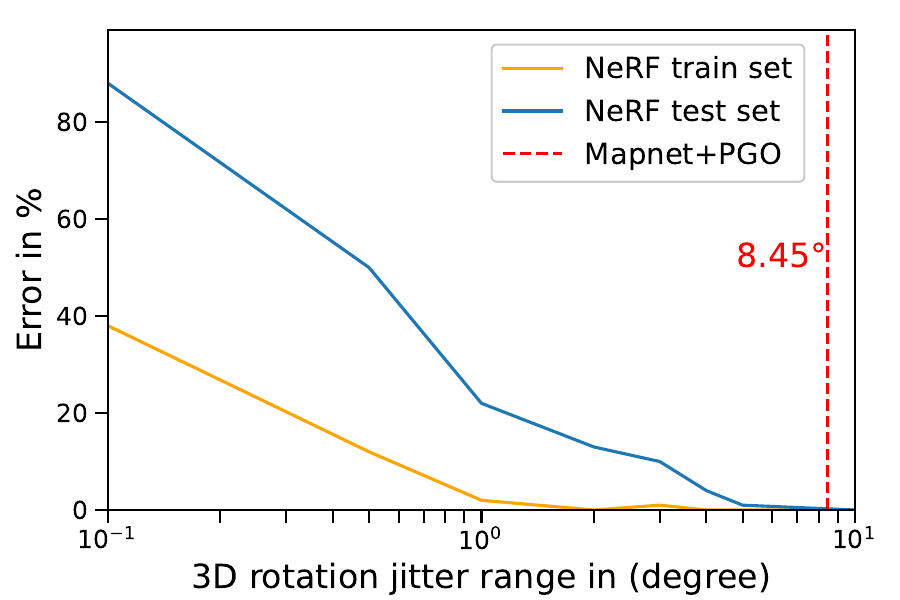}%
        \subcaption{\small 3D rotation jitter only}
    \end{subfigure}
 \caption{We feed randomly perturbed poses to the NeRF model trained in Heads, and validate the robustness of our photometric loss $\mathcal{L}_{photo}$. By cumulatively counting the chances of $\mathcal{L}_{ph\_perturb} < \mathcal{L}_{ph\_GT}$, the \textit{error rate} remains nearly 0 when the perturbation is smaller to the Mapnet+PGO reference threshold, for both translation and rotation. This indicates that the photometric loss from direct matching is effective to supervise the training of a pose regression network.}
 \label{fig:Random_jitter_pose}
 \end{figure}

\begin{table*}[t]
  \centering
  \resizebox{0.7\linewidth}{!}{
      \begin{tabular}{llllll}
         \toprule
         Scene & \makecell[c]{Geo. PoseNet\\~\cite{Kendall17}} & \makecell[c]{PoseNet\\(ResNet34)~\cite{Brahmbhatt18}} & \makecell[c]{PoseNet+logq\\(ResNet34)~\cite{Brahmbhatt18}} & \makecell[c]{PoseNet+SE(3)\\(ResNet34)} & \makecell[c]{PoseNet+SE(3)\\(MobileNetV2)} \\
         \midrule
         \makecell[c]{Backbone Error\\(Top-1/Top-5)} & 31.3\%/11.1\%  & \textbf{26.7\%/8.58\%}  & \textbf{26.7\%/8.58\%} & \textbf{26.7\%/8.58\%} & 28.12\%/9.71\%  \\
         \midrule
         Average     & 0.23m, 8.12\degree  & 0.23m, 8.49\degree  & 0.22m, 8.07\degree  & \textbf{0.21m}, 8.71\degree& \textbf{0.21m}, \textbf{7.84\degree} \\
         \bottomrule
         \\
      \end{tabular}
      
  }
  \caption{A comparison between our SE(3) PoseNet baseline and other quaternion-based baselines. Our direct SE(3) supervision offers competitive results with both backbones while removing the need for balancing rotation and translation terms.}
  \label{table:5}
\end{table*}


Intuitively, the greater range we perturb the ground truth pose with, the more displacement in the appearance of rendered images will have and shall lead to a lower \textit{error rate}. In this experiment, we jitter poses in the range between $\pm{[0.01m, 1m]}$ for 3D translation and the range between $\pm{[0.1\degree, 10\degree]}$ for 3D rotation movement. For each of the selected scales, we randomly jitter 500 poses to estimate the expected \textit{error rate}. As the results are shown in~\cref{fig:Random_jitter_pose}, both 3D translation and 3D rotation \textit{error rates} drop close to 0 below the reference threshold of MapNet+PGO, which may explain why training with direct matching can obtain better performance overall.

\paragraph{Effectiveness of Regressing SO(3)}
We also compares our baseline pose regression model with prior baselines from PoseNet and MapNet in \cref{table:5}. We achieve on-par results by directly replacing the rotation representation from quaternion to SO(3) rotation representation. Our MobileNetV2 performs overall the best in terms of average results. We use identical training hyperparameters to train our baseline model for each scene, and our MobileNetV2 feature extractor is not the best regarding the ImageNet benchmark compared to prior baselines. Our baseline models' superior performance indicates that our rotation representation is just as effective as quaternion representation. A full table on scene specific performance is provided in the supplementary material.

\paragraph{Summary of Ablation}
We justify our design decisions to show how each component variation contributes to the relocalization performance in \cref{table:4}. First, replacing the SE(3) representation with separate quaternion rotation and translation position terms leads to lower accuracy due to the balancing requirement of the two terms during training. The most significant performance drop is when removing the coarse-to-fine training strategy on NeRF. It indicates that the NVS reconstruction quality does affect the overall relocalization accuracy. In addition, we observe that using full NeRF architecture to train our model can obtain slight accuracy improvement. However, this is at the cost of a much longer training time (i.e. 22hrs vs. 78hrs on Kitchen). We observe that the same phenomena hold valid when training with unlabeled data as well.
\begin{table}[h]
   \centering
   \resizebox{0.7\linewidth}{!}{%
   \begin{tabular}{l|l}
   \toprule
   Method & 7 Scenes\\
   \midrule

  
  Direct-PN & 0.20m, 7.26\degree\\
  - SE(3) & 0.21m, 7.58\degree\\
  - Coarse-to-fine & 0.22m, 7.91\degree\\
  - Direct Matching & 0.22m, 8.07\degree\\
  \midrule
  Direct-PN + Full NeRF & \textbf{0.20m, 7.16\degree}\\
   \bottomrule   
   \end{tabular}%
   }
   \caption{A performance breakdown for each component in our method. The performance drops for when our modifications on SE(3), coarse-to-fine encoding, and the direct-matching module are removed. The performance improves slightly if a full-size NeRF \cite{Mildenhall20} (with the hierarchical architecture, and a MLP with deeper and wider layers), but at the cost of a much longer training time.}
   \label{table:4}
\end{table} 

\section{Conclusion and Discussion} \label{sec:conclusion}
In this work, we show that one can use a diﬀerentiable renderer to improve pose regression performance. We present a relocalization pipeline that outperforms previous single-image APR methods on the 7-Scenes benchmark and achieves state-of-the-art performance on the LLFF dataset, with two main contributions. First, we joint a direct matching module with a pose regression network, offering superior performance while maintaining low inference cost. Second, we further boost our method's performance by applying a simple but effective semi-supervised training scheme to unlabeled data. To adapt with outward-looking relocalization datasets, which violates assumptions in NeRF, we employ a coarse-to-fine positional encoding strategy to improve rendering qualities.

One of the limitations of this work is that the effectiveness of our direct matching highly depends on the robustness of the NVS methods, which might fail for various reasons. For example, NeRF does not perform well in large-scale scenes, dynamic environments, or outdoor scenarios where auto-exposure fluctuates. The next challenge for us is to circumvent the assumptions made in NeRF so that our method is extensible to more challenging scenarios.\\

\paragraph{Acknowledgements}
We thank Kejie Li for his advice on experimental design and generous help to polish our paper. We also appreciate Henry Howard-Jenkins and Theo W. Costain for some great comments and discussions.





\clearpage
\section{Supplementary}
\subsection{Implementation Details}
\begin{figure}[!h]
   \centering
   \includegraphics[width=\linewidth]{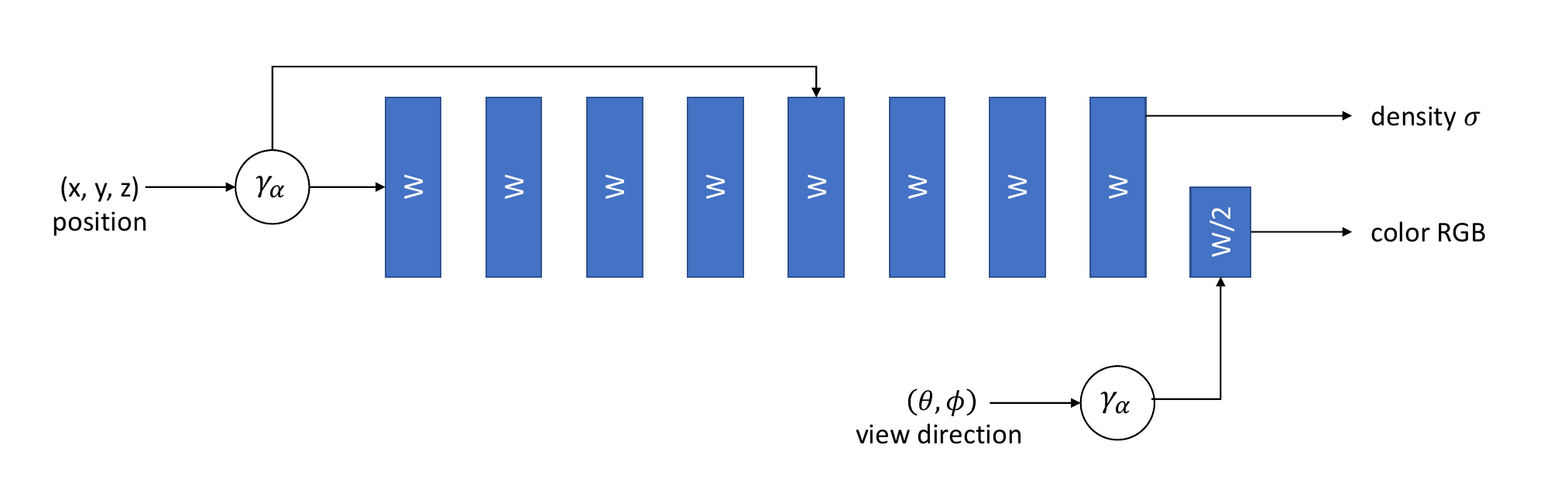}
   \caption{Our NeRF architecture. We adapt a coarse-to-fine encoding approach \cite{park2020nerfies} for both positions and view directions to mitigate artifacts in NeRF reconstruction caused by outward-looking scenes and video distortions. We set $W=128$ in our implementation.}
\label{fig:NeRF_architecture}
\end{figure}

\paragraph{Architectures Details}
The proposed pipeline includes a pose regression model for camera pose predictions and an NVS system for synthesis images. Specifically, we use a modified PoseNet model and a modified NeRF in our experiments. We summarize the details of the pose regression network architecture in~\cref{table:6}. The architecture of our NeRF~(\cref{fig:NeRF_architecture}) mainly follows the original implementation of Mildenhall \etal~\cite{Mildenhall20}, except we apply a coarse-to-fine positional encoding $\gamma_{\alpha}(p)$ for both positions and directions, and we only use a coarse model with 128 samples along each ray. The entire implementation is written in PyTorch, and the NeRF code is built upon an open-sourced repository nerf-pytorch ~\cite{PyTorchNeRF}.

\paragraph{Orthogonalize the Rotation Matrix} 
As we have mentioned in the paper, the ground truth supervision $\mathcal{L}_{gt}$ is an approximation of the correct geometric loss. The regressed rotation matrix $\hat{\mathbf{R}}$ is not guaranteed to be in SO(3) manifold. To solve this, we apply a singular value decomposition (SVD) operation during testing:

\begin{equation}
    \text{SVD}(\hat{\mathbf{R}}) = \mathbf{U} \mathbf{\Sigma} \mathbf{V}^T,
\end{equation}
\begin{equation}
    \mathbf{\hat{R}_o} = \mathbf{U} \mathbf{V}^T,
\end{equation}
where $\mathbf{\hat{R}_o}$ denotes the orthogonalized rotation matrix.

\begin{table}[!ht]
   \centering
   \resizebox{0.7\linewidth}{!}{%
   \begin{tabular}{c|c|c|c|c|c}
   \toprule
   Input & Operator & $t$ & $m$ & $n$ & $s$\\
   \midrule
   $240 \times 320 \times 3$ & conv2d & - & 32 & 1 & 2\\
   \midrule
   $120 \times 160 \times 32$ & bottleneck & 1 & 16 & 1 & 1\\
   \midrule
   $120 \times 160 \times 16$ & bottleneck & 6 & 24 & 2 & 2\\
   \midrule
   $60 \times 80 \times 24$ & bottleneck & 6 & 32 & 3 & 2\\
   \midrule
   $30 \times 40 \times 32$ & bottleneck & 6 & 64 & 4 & 2\\
   \midrule
   $15 \times 20 \times 64$ & bottleneck & 6 & 96 & 3 & 1\\
   \midrule
   $15 \times 20 \times 96$ & bottleneck & 6 & 160 & 3 & 2\\
   \midrule
   $8 \times 10 \times 160$ & bottleneck & 6 & 320 & 1 & 1\\
   \midrule
   $8 \times 10 \times 320$ & conv2d $1\times1$ & - & 1280 & 1 & 1\\
   \midrule
   $8 \times 10 \times 1280$ & avgpool & - & - & 1 & -\\
   \midrule
   $1 \times 1 \times 1280$ & fc & - & 12 & 1 & -\\
   \bottomrule   
   \end{tabular}%
   }
   \caption{Baseline pose regression network architecture of Direct-PoseNet, using an input image size $240\times320\times3$ as an example. The backbone is MobileNetV2~\cite{MobileNetv2}, with $n$ repeated times for each operator and $m$ output channels. $s$ represents the stride and $t$ represents the expansion factor.}
   \label{table:6}
\end{table}

\begin{figure}
    \centering
    \includegraphics[width=1\linewidth]{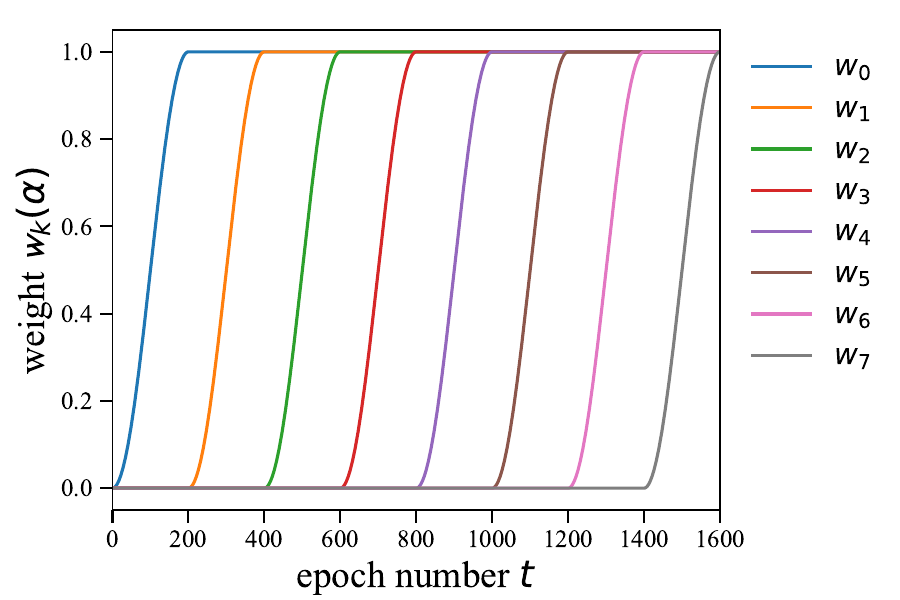}%
    \caption{An example of $w_k$ activation used in the main paper. The number of frequency band being activated increase as epoch iterations increase.}
\label{fig:w_k2}
\end{figure}

\paragraph{Positional Encoding} 
Strong artifacts occur in synthetic images if the NeRF is applied in the relocalization task without modification. This phenomenon appears because: a) NeRF is not designed for outward-looking scenes and b) photometric consistency is violated when auto-focus/exposure fluctuation and rolling shutter effect appear. As we addressed in the main paper, we mitigate this issue by adapting a coarse-to-fine positional encoding strategy $\gamma_{\alpha}(p)$ proposed by Nerfies~\cite{park2020nerfies}, and illustrate this strategy in \cref{fig:w_k2}. Specifically, an input signal $p$ is encoded by
\begin{equation}
    \begin{aligned}
    \gamma_{\alpha}(p) = {} &[p, \dots, w_k(\alpha_t) \sin (2^{k} \pi p), \\
    & w_k(\alpha_t) \cos (2^{k} \pi p), \dots ],
    \end{aligned}
\end{equation}
where $0 \leq k \leq m-1, m\in\mathbb{N}$ and $w_{k}(\alpha_t)$ activates each band over epoch $t$, controlled by $\alpha_t = mt/N$. We denote $N$ as the maximum epoch number in training and the weight $w_{k}(\alpha_t)$ is defined as:
\begin{align}
    w_{k}(\alpha_t)=\frac{(1-\cos (\pi \operatorname{clamp}(\alpha_t-k, 0,1))}{2}.
\end{align}

\begin{table}[!htb]
  \centering
  \resizebox{\linewidth}{!}{
  \begin{tabular}{c|ccc|cc}
  \toprule
  \multicolumn{1}{c|}{} &
  \multicolumn{3}{c|}{without unlabeled data} &
         \multicolumn{2}{c}{with unlabeled data} \\
  \midrule
  Model & \makecell[c]{PoseNet+logq~\cite{Brahmbhatt18}} & \makecell[c]{MapNet~\cite{Brahmbhatt18}} & \makecell[c]{Direct-PN} & \makecell[c]{MapNet+PGO~\cite{Brahmbhatt18}} & \makecell[c]{Direct-PN+U}\\
  \midrule
  Avg. Median & \makecell[c]{0.23m, 8.49\degree}     & \makecell[c]{0.21m, 7.77\degree} & \makecell[c]{\textbf{0.20m, 7.26\degree}} & \makecell[c]{0.18m, 6.55\degree} & \makecell[c]{\textbf{0.16m, 5.17\degree}}\\
  \midrule
  Avg. Mean & \makecell[c]{0.28m, 10.43\degree}     & \makecell[c]{0.27m, 10.08\degree} & \makecell[c]{\textbf{0.25m, 8.98\degree}} & \makecell[c]{0.22m, 7.89\degree} & \makecell[c]{\textbf{0.21m, 7.02\degree}}\\
  \bottomrule   
  \end{tabular}%
  }
  \caption{A comparison of average median errors and average mean errors on the 7-Scenes dataset.}
  \label{table:7}
\end{table}



\subsection{Additional Ablation Study}
\paragraph{More Results on the 7-Scenes Dataset}
We further compare our method with prior state-of-the-art methods on the 7-Scenes dataset (\cref{table:7}), showing our pipeline outperforms them in average median errors and average mean errors.
\begin{table}
   \centering
   \resizebox{\linewidth}{!}{
      \begin{tabular}{llllll}
         \toprule
         Scene & \makecell[c]{Geo. PoseNet\\~\cite{Kendall17}} & \makecell[c]{PoseNet\\(ResNet34)~\cite{Brahmbhatt18}} & \makecell[c]{PoseNet+logq\\(ResNet34)~\cite{Brahmbhatt18}} & \makecell[c]{PoseNet+SE(3)\\(ResNet34)} & \makecell[c]{PoseNet+SE(3)\\(MobileNetV2)} \\
         \midrule
         \makecell[c]{Backbone Error\\(Top-1/Top-5)} & 31.3\%/11.1\%  & \textbf{26.7\%/8.58\%}  & \textbf{26.7\%/8.58\%} & \textbf{26.7\%/8.58\%} & 28.12\%/9.71\%  \\
         \midrule
         Chess       & 0.13m, 4.48\degree  & \textbf{0.11m}, 4.24\degree  & \textbf{0.11m}, 4.29\degree  & \textbf{0.11m}, 4.53\degree & \textbf{0.11m}, \textbf{3.95\degree} \\
         Fire        & \textbf{0.27m}, 11.30\degree & 0.29m, 11.68\degree & \textbf{0.27m}, 12.13\degree & 0.28m, 11.65\degree         & \textbf{0.27m}, \textbf{10.15\degree} \\
         Heads       & \textbf{0.17m}, 13.00\degree & 0.20m, 13.11\degree & 0.19m, \textbf{12.15\degree} & \textbf{0.17m}, 13.76\degree & \textbf{0.17m}, 13.30\degree \\
         Office      & 0.19m, \textbf{5.55\degree}  & 0.19m, 6.40\degree  & 0.19m, 6.35\degree  & 0.18m, 5.92\degree                    & \textbf{0.17m}, 6.25\degree \\
         Pumpkin     & 0.26m, 4.75\degree  & 0.23m, 5.77\degree  & 0.22m, 5.05\degree  & \textbf{0.20m}, 6.11\degree           & 0.22m, \textbf{4.58\degree} \\
         Kitchen     & \textbf{0.23m}, 5.35\degree  & 0.27m, 5.81\degree  & 0.25m, \textbf{5.27\degree}  & 0.24m, 6.22\degree           & 0.24m, 5.47\degree \\
         Stairs      & 0.35m, 12.40\degree & 0.31m. 12.43\degree & 0.30m, 11.29\degree & \textbf{0.29m}, 12.76\degree& 0.30m, \textbf{11.20\degree} \\
         \midrule
         Average     & 0.23m, 8.12\degree  & 0.23m, 8.49\degree  & 0.22m, 8.07\degree  & \textbf{0.21m}, 8.71\degree& \textbf{0.21m}, \textbf{7.84\degree} \\
         \bottomrule
         \\
      \end{tabular}
      
   }
   \caption{A per scene based comparison between our SE(3) PoseNet baseline and other quaternion-based baselines, evaluated with median translation and rotation error on the 7-Scenes. Two columns to the right are results with our direct SE(3) supervision.}
   \label{table:8}
\end{table}
\cref{table:8} shows the scene-specific comparison between SO(3) and quaternion-based representation. The quaternion-based results were provided by \cite{Kendall17, Brahmbhatt18}, and their performances are confirmed based on their released code. The quaternion-based models were trained using geometric consistency loss \cite{Brahmbhatt18}, and the SE(3) models were trained using L2 loss without balancing translation and rotation terms.

\paragraph{About $\lambda$s}
The reconstruction loss tends to have an order of magnitude larger value than the pose loss. In our paper, we didn't heavily tune the $\lambda$s. We experimentally pull both losses into closer scales. Even with our default $\lambda_1=0.3, \lambda_2=0.7$ values, we observe that the weighted reconstruction loss is usually the dominant term of the combined loss, which proves the benefits of our architecture. Table \ref{table:1} shows our experiments on 3 of 7-scenes with different $\lambda$ settings. Although the result seems mixed, we argue both pose loss and photometric loss are important. For scenes with low texture or flat background, i.e., Lego scene from NeRF Synthetic dataset \cite{Mildenhall20}, pose loss ensures the regressed pose is regularized in relevant positions.
\begin{table}[h]
   \resizebox{\linewidth}{!}{%
   \begin{tabular}{l|l|lllll}
   \toprule
   Dataset & Scene & \makecell[c]{$\lambda_1=0.1$\\$\lambda_2=0.9$} 
  & \makecell[c]{$\lambda_1=0.3$\\$\lambda_2=0.7$} & \makecell[c]{$\lambda_1=0.5$\\$\lambda_2=0.5$}
  & \makecell[c]{$\lambda_1=0.7$\\$\lambda_2=0.3$} & \makecell[c]{$\lambda_1=0.9$\\$\lambda_2=0.1$}\\
   \midrule
   7 Scenes & Heads & \makecell[c]{\textbf{0.17}, 13.08\degree} & \makecell[c]{\textbf{0.17}, 13.1\degree} & \makecell[c]{\textbf{0.17}, \textbf{12.87\degree}} & \makecell[c]{\textbf{0.17}, 13.1\degree} & \makecell[c]{\textbf{0.17}, 13.26\degree}\\ 
   
   7 Scenes & Fire & \makecell[c]{0.28, \textbf{8.48\degree}} & \makecell[c]{\textbf{0.27}, 8.66\degree} & \makecell[c]{\textbf{0.27}, 8.61\degree} & \makecell[c]{\textbf{0.27}, 8.87\degree} & \makecell[c]{\textbf{0.27}, 9.36\degree}\\
   
   7 Scenes & Pumpkin & \makecell[c]{\textbf{0.19},\textbf{3.60\degree}} & \makecell[c]{\textbf{0.19},3.85\degree} & \makecell[c]{\textbf{0.19},3.77\degree} & \makecell[c]{0.20,3.68\degree} & \makecell[c]{\textbf{0.19}, 3.64\degree}\\
   \midrule
   
  \makecell[l]{NeRF\\Synthetic} & Lego & \makecell[c]{0.167, 2.9\degree} & \makecell[c]{\textbf{0.117}, \textbf{2.7\degree}} & \makecell[c]{0.182, 4.7\degree} & \makecell[c]{0.194, 5.1\degree} & \makecell[c]{0.276, 5.8\degree}\\
   \bottomrule   
   \end{tabular}%
   }
   \caption{Result of using different $\lambda_1$ and $\lambda_2$ values on Heads, Fire, and Pumpkin in 7-Scenes dataset (first 3 rows). We also tested in Lego scene (bottom row) on the NeRF synthesis datasets \cite{Mildenhall20}, which contains large areas of textureless background.}
   \label{table:1}
   \vspace{-5mm}
\end{table}



\clearpage
{\small
\bibliographystyle{toolkit/ieee_fullname}
\bibliography{main}
}

\end{document}